\documentclass[10pt,twocolumn,letterpaper]{article}

\usepackage{iccv}
\usepackage{times}
\usepackage{epsfig}
\usepackage{graphicx}
\usepackage{amsmath}
\usepackage{amssymb}
\usepackage{CJKutf8}
\usepackage{mathrsfs}
\usepackage{subcaption}
\usepackage{makecell}
\usepackage{float}
\usepackage{dsfont}
\usepackage{booktabs}
\usepackage{setspace}

\usepackage[breaklinks=true,bookmarks=false]{hyperref}

\iccvfinalcopy 

\def\iccvPaperID{3391} 

\ificcvfinal\fi

\begin{document}

\title{Lifelong GAN: Continual Learning for Conditional Image Generation}

\author{Mengyao Zhai\textsuperscript{1,2}\thanks{Equal Contribution} \,, Lei Chen\textsuperscript{1,2}\footnotemark[1] \,, Fred Tung\textsuperscript{1,2}, Jiawei He\textsuperscript{1,2}, Megha Nawhal\textsuperscript{1,2}, Greg Mori\textsuperscript{1,2}\\
\textsuperscript{1}Simon Fraser University   \;\;\;      \textsuperscript{2}Borealis AI\\
{\tt\small \{mzhai, chenleic, ftung, jha203, mnawhal\}@sfu.ca \; mori@cs.sfu.ca}\vspace{-1cm}}


\makeatletter
\vspace{-1cm}
\g@addto@macro\@maketitle{
  \begin{figure}[H]
  \setlength{\linewidth}{\textwidth}
  \setlength{\hsize}{\textwidth}
  \centering
  \includegraphics[width=0.98\textwidth]{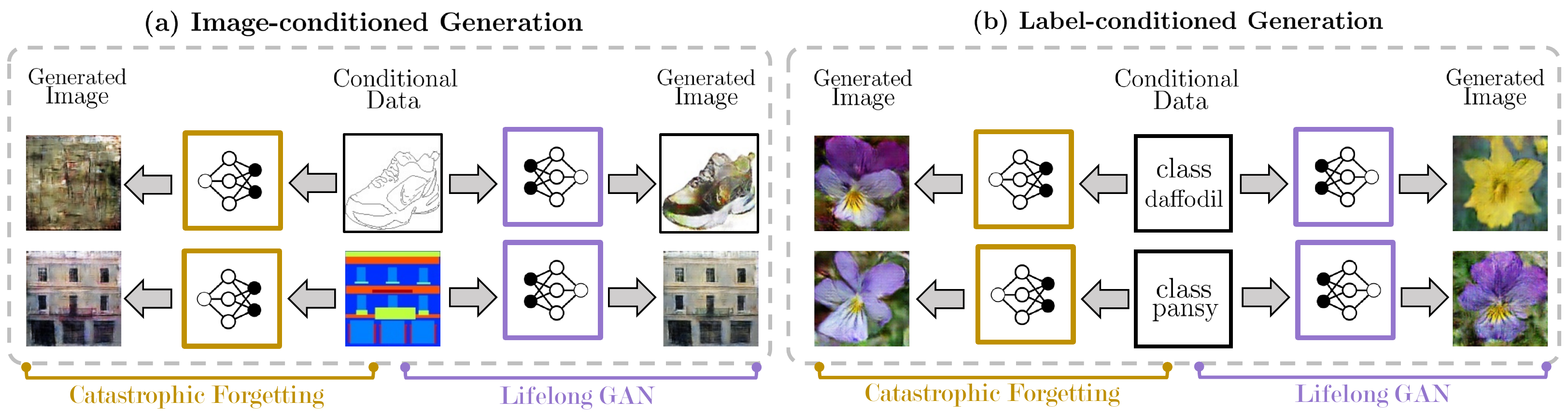}
  \vspace{-0.05in}
  \caption{\textbf{Lifelong learning of conditional image generation.} Traditional training methods suffer from catastrophic forgetting: when we add new tasks, the network forgets how to perform previous tasks. Our Lifelong GAN is a generic framework for conditional image generation that applies to various types of conditional inputs (\eg labels and images).}
  \label{fig:introduction_figure}
  \end{figure}
}
\makeatother

\maketitle
\ificcvfinal\thispagestyle{empty}\fi
\begin{abstract}
Lifelong learning is challenging for deep neural networks due to their susceptibility to catastrophic forgetting. Catastrophic forgetting occurs when a trained network is not able to maintain its ability to accomplish previously learned tasks when it is trained to perform new tasks. We study the problem of lifelong learning for generative models, extending a trained network to new conditional generation tasks without forgetting previous tasks, while assuming access to the training data for the current task only. In contrast to state-of-the-art  memory replay based approaches which are limited to label-conditioned image generation tasks, a more generic framework for continual learning of generative models under different conditional image generation settings is proposed in this paper. Lifelong GAN employs knowledge distillation to transfer learned knowledge from previous networks to the new network. This makes it possible to perform image-conditioned generation tasks in a lifelong learning setting. We validate Lifelong GAN for both image-conditioned and label-conditioned generation tasks, and provide qualitative and quantitative results to show the generality and effectiveness of our method.
\end{abstract}


\section{Introduction}

Learning is a lifelong process for humans. We acquire knowledge throughout our lives so that we become more efficient and versatile facing new tasks. The accumulation of knowledge in turn accelerates our acquisition of new skills. In contrast to human learning, lifelong learning remains an open challenge for modern deep learning systems. It is well known that deep neural networks are susceptible to a phenomenon known as \textit{catastrophic forgetting}~\cite{mccloskey1989catastrophic}. Catastrophic forgetting occurs when a trained neural network is not able to maintain its ability to accomplish previously learned tasks when it is adapted to perform new tasks. 

Consider the example in Figure~\ref{fig:introduction_figure}. A generative model is first trained on the task $\mathrm{edges} \rightarrow \mathrm{shoes}$. Given a new task $\mathrm{segmentations} \rightarrow \mathrm{facades}$, a new model is initialized from the previous one and fine-tuned for the new task. After training, the model forgets about the previous task and cannot generate shoe images given edge images as inputs. One way to address this would be to combine the training data for the current task with the training data for all previous tasks and then train the model using the joint data. Unfortunately, this approach is not scalable in general: as new tasks are added, the storage requirements and training time of the joint data grow without bound. In addition, the models for previous tasks may be trained using private or privileged data which is not accessible during the training of the current task. The challenge in lifelong learning is therefore to extend the model to accomplish the current task, without forgetting how to accomplish previous tasks in scenarios where we are restricted to the training data for only the current task. In this work, we work under the assumption that we only have access to a model trained on previous tasks without access to the previous data.

Recent efforts~\cite{shmelkov2017incremental, castro2018end, churchill2012practice} have demonstrated how discriminative models could be incrementally learnt for a sequence of tasks. Despite the success of these efforts, lifelong learning in generative settings remains an open problem. Parameter regularization~\cite{seff2017continual,kirkpatrick2017overcoming} has been adapted from discriminative models to generative models, but poor performance is observed~\cite{wu2018memory}.
The state-of-the-art continual learning generative frameworks~\cite{seff2017continual,wu2018memory} are built on \textit{memory replay} which treats generated data from previous tasks as part of the training examples in the new tasks. Although memory replay has been shown to alleviate the catastrophic forgetting problem by taking advantage of the generative setting, its applicability is limited to label-conditioned generation tasks. In particular, replay based methods cannot be extended to image-conditioned generation. The reason lies in that no conditional image can be accessed to generate replay training pairs for previous tasks. Therefore, a more generic continual learning framework that can enable various conditional generation tasks is valuable.

In this paper, we introduce a generic continual learning framework \textit{Lifelong GAN} that can be applied to both image-conditioned and label-conditioned image generation. We employ \textit{knowledge distillation}~\cite{hinton2015distilling} to address catastrophic forgetting for conditional generative continual learning tasks. Given a new task, Lifelong GAN learns to perform this task, and to keep the memory of previous tasks, information is extracted from a previously trained network and distilled to the new network during training by encouraging the two networks to produce similar output values or visual patterns. To the best of our knowledge, we are the first to utilize the principle of knowledge distillation for continual learning generative frameworks.

To summarize, our contributions are as follows. \textit{First}, we propose a generic framework for continual learning of conditional image generation models. \textit{Second}, we validate the effectiveness of our approach for two different types of conditional inputs: (1) image-conditioned generation, and (2) label-conditioned generation, and provide qualitative and quantitative results to illustrate the capability of our GAN framework to learn new generation tasks without the catastrophic forgetting of previous tasks. \textit{Third}, we illustrate the generality of our framework by performing continual learning across diverse data domains.

\section{Related Work}
\noindent
\textbf{Conditional GANs.}
Image generation has achieved great success since the introduction of GANs~\cite{goodfellow2014GAN}. There also has been rapid progress in the field of conditional image generation~\cite{mirza2014conditional}. Conditional image generation tasks can be typically categorized as image-conditioned image generation and label-conditioned image generation.

Recent image-conditioned models have shown promising results for numerous image-to-image translation tasks such as maps $\rightarrow$ satellite images, sketches $\rightarrow$ photos, labels $\rightarrow$ images~\cite{isola2016pix2pix, zhu2017toward,zhu2017unpaired}, future frame prediction ~\cite{Villegas2017LearningToGenerate}, superresolution~\cite{ledig2017photo}, and inpainting~\cite{yeh2017semantic}. Moreover, images can be stylized by disentangling the style and the content~\cite{johnson2016perceptual,luan2017deep} or by encoding styles into a stylebank (set of convolution filters)~\cite{chen2017stylebank}. Models~\cite{ZhaiDCCDM18,ma2017pose} for rendering a person's appearance onto a given pose have shown to be effective for person re-identification.
Label-conditioned models~\cite{chen2016infogan,chongxuan2017triple} have also been explored for generating images for specific categories. 

\vspace{0.05in}
\noindent
\textbf{Knowledge Distillation.} Proposed by Hinton et al.~\cite{hinton2015distilling}, knowledge distillation is designed for transferring knowledge from a teacher classifier to a student classifier. The teacher classifier normally would have more privileged information~\cite{vapnik2015learning} compared with the student classifier. The privileged information includes two aspects. The first aspect is referred to as the learning power, namely the size of the neural networks.  A student classifier could have a more compact network structure compared with the teacher classifier, and by distilling knowledge from the teacher classifier to student classifier, the student classifier would have similar or even better classification performance than the teacher network.  Relevant applications include network compression~\cite{polino2018model} and network training acceleration~\cite{wang2018kdgan}. The second aspect is the learning resources, namely the amount of input data. The teacher classifier could have more learning resources and see more data that the student cannot see. 
Compared with the first aspect, this aspect is relatively unexplored and is the focus of our work.

\vspace{0.05in}
\noindent
\textbf{Continual Learning.}
For discriminative tasks e.g.\ classification, many works have been proposed recently for solving the problem of catastrophic forgetting in computer vision~\cite{shmelkov2017incremental,castro2018end} and robotics~\cite{churchill2012practice}. Shmelkov \etal~\cite{shmelkov2017incremental}, Castro \etal~\cite{castro2018end} and Li \etal~\cite{li2017learning} employed a distillation loss that measures the discrepancy between the output of the old and new network. Serr{\`a} \etal~\cite{Serr2018OvercomingCF} proposed a task-based hard attention mechanism to learn new tasks without forgetting previous tasks. \textit{EWC}~\cite{kirkpatrick2017overcoming}, \textit{RWALK}~\cite{chaudhry2018riemannian} and \textit{MAS}~\cite{aljundi2018memory} are regularization-based approaches which regularize the network parameters when learning new tasks. \textit{GEM} based approaches~\cite{lopez2017gradient,chaudhry2018efficient} store part of the training data from previous tasks to regularize the gradients when learning new tasks and aim at better performance in the single pass setting.

For generative tasks, relatively less work studies the problem of catastrophic forgetting. Continual generative modeling was first introduced by Seff \etal~\cite{seff2017continual}. Their approach incorporated the idea of \textit{EWC} into the loss function of GANs. The idea of \textit{memory replay}, also mentioned in~\cite{seff2017continual}, is well explored by Wu \etal~\cite{wu2018memory} for label-conditioned image generation. Approaches based on \textit{EWC} have been explored for the task of label-conditioned image generation~\cite{seff2017continual,wu2018memory} to generate more realistic images, but they present limited capability in both remembering previous categories and generating high quality images.

In this paper, we introduce knowledge distillation within continual generative model learning, which has not been explored before. Our approach can be applied to both image-conditioned generation, for which the replay mechanism is not applicable, and label-conditioned image generation.


\section{Approach}

\begin{figure*}[h]
    \centering
    \includegraphics[width=0.9\textwidth]{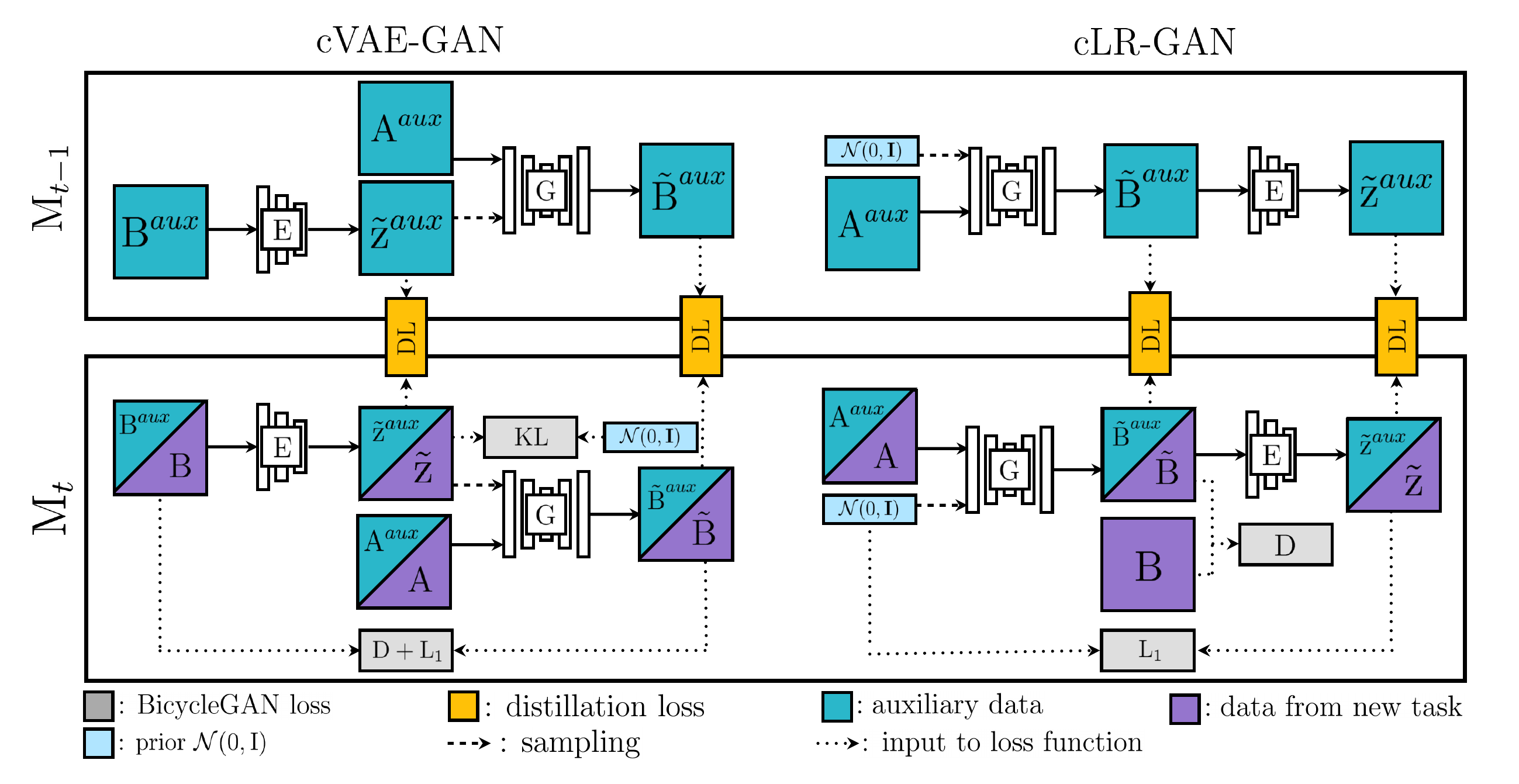}
    \vspace{-0.3cm}
    \caption{\textbf{Overview of Lifelong GAN}. Given training data for the $t^{th}$ task, model $M_{t}$ is trained to learn this current task. To avoid forgetting previous tasks, knowledge distillation is adopted to distill information from model $M_{t-1}$ to model $M_{t}$ by encouraging the two networks to produce similar output values or patterns given the auxiliary data as inputs.}
    \vspace{-0.5cm}
    \label{fig:overview}
\end{figure*}


Our proposed Lifelong GAN addresses catastrophic forgetting using knowledge distillation and, in contrast to replay based methods, can be applied to continually learn both label-conditioned and image-conditioned generation tasks. In this paper, we build our model on the state-of-the-art BicycleGAN~\cite{zhu2017toward} model. Our overall approach for continual learning for a generative model is illustrated in Figure~\ref{fig:overview}. Given data from the current task, Lifelong GAN learns to perform this task, and to keep the memory of previous tasks, knowledge distillation is adopted to distill information from a previously trained network to the current network by encouraging the two networks to produce similar output values or patterns given the same input. To avoid ``conflicts" that arise when having two desired outputs (current training goal and outputs from previous model) given the same input, we generate auxiliary data for distillation from the current data via two operations \textit{Montage} and \textit{Swap}.

\subsection{Background: BicycleGAN}
\label{sec:background}
We first introduce the state-of-the-art BicycleGAN~\cite{zhu2017toward} on which our model is built. Let the encoder be $E$, generator be $G$ and discriminator be $D$. Denote the training set as $\mathbb{S} = \{(\mathbf{A}^i,\mathbf{B}^i)| \mathbf{A}^i \in \mathbb{A}, \mathbf{B}^i \in \mathbb{B}\}$ where $\mathbb{A}$ and $\mathbb{B}$ stand for the set of conditional and ground-truth images. For simplicity, we use the notations $\mathbf{A}, \mathbf{B}$ for an instance from the respective domain. The Bicycle-GAN model consists of two cycles and resembles two GAN models: cVAE-GAN and cLR-GAN. Now, we describe the two cycles in detail.

\vspace{0.05in}
\noindent
\textbf{cVAE-GAN.} 
The first model is \textit{cVAE-GAN}, which first encodes ground truth image $\mathbf{B}$ to latent code $\mathbf{\widetilde{z}}$ using the encoder $E$, then reconstructs the ground truth image as $\mathbf{\widetilde{B}}$ given the conditional image $\mathbf{A}$ and encoded latent code $\mathbf{\widetilde{z}}$. 

The loss of \textit{cVAE-GAN} consists of three terms: $\mathcal{L}_{1}^{\mathrm{image}}=\mathds{E}_{\mathbf{A,B} \sim p(\mathbf{A,B}),\mathbf{\widetilde{z}} \sim E(\mathbf{B})}[||\mathbf{B} - G(\mathbf{A},\mathbf{\widetilde{z}})||_{1}]$ which encourages the output of the generator to match the input; $\mathcal{L}_{\mathrm{KL}}=\mathds{E}_{\mathbf{B} \sim p(\mathbf{B})} [\mathrm{KL}(E(\mathbf{B}) || \mathcal{N}(0,\mathbf{I}))]$ which encourages the encoded latent distribution to be close to a standard Gaussian to enable sampling at inference time; and  $\mathcal{L}_{\mathrm{GAN}}^{\mathrm{cVAE}}$, the standard adversarial loss which encourages the generator to generate images that are not distinguishable from real images by the discriminator. The objective function of the \textit{cVAE-GAN} is:
\begin{equation}
\begin{split}
    \mathcal{L}_{\mathrm{cVAE-GAN}} = \underset{G,E}{\min}\,\underset{D}{\max}\quad  \mathcal{L}_{\mathrm{GAN}}^{\mathrm{cVAE}} + \lambda \mathcal{L}_{1}^{\mathrm{image}} + \lambda_{\mathrm{KL}} \mathcal{L}_{\mathrm{KL}},
\end{split}
\end{equation}
where $\lambda$ and $\lambda_{\mathrm{KL}}$ are loss weights for encoding and image reconstruction, respectively.

\vspace{0.05in}
\noindent
\textbf{cLR-GAN.} 
The second model is \textit{cLR-GAN}, which first generates a image $\mathbf{\widetilde{B}}$ given the conditional data $\mathbf{A}$ and latent code $\mathbf{z}$, then reconstructs the latent code as $\mathbf{\widetilde{z}}$ to enforce the latent code $\mathbf{z}$ is used. 

The loss of \textit{cLR-GAN} consists of two terms: $\mathcal{L}_{1}^{\mathrm{latent}}=\mathds{E}_{\mathbf{A} \sim p(\mathbf{A}),\mathbf{z} \sim p(\mathbf{z})}[||\mathbf{z} - E(G(\mathbf{A},\mathbf{z}))||_{1}]$ which encourages utilization of the latent code via reconstruction; and $\mathcal{L}_{\mathrm{GAN}}^{\mathrm{cLR}}$, the standard adversarial loss which encourages the generator to generate images that are not distinguishable from real images by the discriminator. The objective function of the \textit{cLR-GAN} is:
\begin{equation}
\begin{split}
\mathcal{L}_{\mathrm{cLR-GAN}} = \underset{G,E}{\min}\,\underset{D}{\max}\quad  \mathcal{L}_{\mathrm{GAN}}^{\mathrm{cLR}} + \lambda_{\mathrm{latent}} \mathcal{L}_{1}^{\mathrm{latent}},
\end{split}
\end{equation}
where $\lambda_{\mathrm{latent}}$ is the loss weight for recovering the latent code.

BicycleGAN is proposed to take advantage of both cycles, hence the objective function is:
\begin{equation}
\begin{split}
    \mathcal{L}_{\mathrm{BicycleGAN}} = \underset{G,E}{\min}\,\underset{D}{\max}\quad  
    \mathcal{L}_{\mathrm{cVAE-GAN}} + \mathcal{L}_{\mathrm{cLR-GAN}}.
\end{split}
\end{equation}

\subsection{Lifelong GAN with Knowledge Distillation}
\label{sec:i-kd}

To perform continual learning of conditional generation tasks, the proposed Lifelong GAN is built on top of Bicycle GAN with the adoption of knowledge distillation.
We first introduce the problem formulation, followed by a detailed description of our model, then discuss our strategy to tackle the  conflicting objectives in training. 

\vspace{0.05in}
\noindent
\textbf{Problem Formulation.}
During training of the $t^{th}$ task, we are given a dataset of $N_t$ paired instances $\mathbb{S}_t=\{(\mathbf{A}_{i,t}, \mathbf{B}_{i,t})|\mathbf{A}_{i,t} \in \mathbb{A}_t, \mathbf{B}_{i,t}\in{}\mathbb{B}_t\}_{i=1}^{N_t}$ where $\mathbb{A}_t$ and $\mathbb{B}_t$ denote the domain of conditional images and ground truth images respectively. For simplicity, we use the notations $\mathbf{A}_t, \mathbf{B}_t$ for an instance from the respective domain. The goal is to train a model $M_{t}$ which can generate images of current task $\mathbf{\widetilde{B}}_{t} \leftarrow (\mathbf{A}_{t},\mathbf{z})$, without forgetting how to generate images of previous tasks $\mathbf{\widetilde{B}}_i \leftarrow (\mathbf{A}_i,\mathbf{z})$,  $i=1,2,...,(t-1)$. 

Let $M_{t}$ be the $t^{th}$ model trained, and $M_{t-1}$ be the $(t-1)^{th}$ model trained. Both $M_{t-1}$ and $M_{t}$ contain two cycles (cVAE-GAN and cLR-GAN) as described in Section~\ref{sec:background}. Inspired by continual learning methods for discriminative models, we prevent the current model $M_{t}$ from forgetting the knowledge learned by the previous model $M_{t-1}$ by inputting the data of the current task $\mathbb{S}_{t}$ to both $M_{t}$ and $M_{t-1}$, and distilling the knowledge from $M_{t-1}$ to $M_{t}$ by encouraging the outputs of $M_{t-1}$ and $M_{t}$ to be similar. We describe the process of knowledge distillation for both cycles as follows.

\vspace{0.05in}
\noindent
\textbf{cVAE-GAN.} Recall from Section \ref{sec:background} that cVAE-GAN has two outputs: the encoded latent code $\mathbf{\widetilde{z}}$ and the reconstructed ground truth image $\mathbf{\widetilde{B}}$. Given ground truth image $\mathbf{B}_{t}$, the encoders $E_{t}$ and $E_{t-1}$ are encouraged to encode it in the same way and produce the same output; given encoded latent code $\mathbf{\widetilde{z}}$ and conditional image $\mathbf{A}_{t}$, the generators $G_{t}$ and $G_{t-1}$ are encouraged to reconstruct the ground truth images in the same way. Therefore, we define the loss for the \textit{cVAE-GAN} cycle with knowledge distillation as: 
\begin{equation}
\begin{split}
    & \mathcal{L}_{\mathrm{cVAE-DL}}^{t} = \mathcal{L}_{\mathrm{cVAE-GAN}}^{t} \\
    & + \beta \mathds{E}_{\mathbf{A}_{t},\mathbf{B}_{t} \sim p(\mathbf{A}_{t},\mathbf{B}_{t})} \,  [||E_{t}(\mathbf{B}_{t})-E_{t-1}(\mathbf{B}_{t})||_{1}\\
    & + ||G_{t}(\mathbf{A}_{t},E_{t}(\mathbf{B}_{t}))-G_{t-1}(\mathbf{A}_{t},E_{t-1}(\mathbf{B}_{t}))||_{1}],
\end{split}
\label{eqn:c1-conflict}
\end{equation}
where $\beta$ is the loss weight for knowledge distillation. 

\vspace{0.05in}
\noindent
\textbf{cLR-GAN.} Recall from Section \ref{sec:background} that cLR-GAN also has two outputs: the generated image $\mathbf{\widetilde{B}}$ and the reconstructed latent code $\mathbf{\widetilde{z}}$. Given the latent code $\mathbf{z}$ and conditional image $\mathbf{A}_{t}$, the generators $G_{t}$ and $G_{t-1}$ are encouraged to generate images in the same way; given the generated image $\mathbf{\widetilde{B}}_{t}$, the encoders $E_{t}$ and $E_{t-1}$ are encouraged to encode the generated images in the same way. Therefore, we define the loss for the \textit{cLR-GAN} cycle as:
\begin{equation}
\begin{split}
    & \mathcal{L}_{\mathrm{cLR-DL}}^{t}  = \mathcal{L}_{\mathrm{cLR-GAN}}^{t} \\
    & + \beta \mathds{E}_{\mathbf{A}_{t} \sim p(\mathbf{A}_{t}),\mathbf{z} \sim p(\mathbf{z})} \, [ ||G_{t}(\mathbf{A}_{t},\mathbf{z})-G_{t-1}(\mathbf{A}_{t},\mathbf{z})||_{1} \\
    & + ||E_{t}(G_{t}(\mathbf{A}_{t},\mathbf{z}))-E_{t-1}(G_{t-1}(\mathbf{A}_{t},\mathbf{z}))||_{1}].\\
\end{split}
\label{eqn:c2-conflict}
\end{equation}

The distillation losses can be defined in several ways, e.g.\ the $L_2$ loss~\cite{ba2014deep,shmelkov2017incremental}, $\mathrm{KL}$ divergence~\cite{hinton2015distilling} or cross-entropy~\cite{hinton2015distilling,castro2018end}. In our approach, we use $L_1$ instead of $L_2$ to avoid blurriness in the generated images. 

Lifelong GAN is proposed to adopt knowledge distillation in both cycles, hence the overall loss function is:
\begin{equation}
\begin{split}
    \mathcal{L}^{t}_{\mathrm{Lifelong-GAN}} = \mathcal{L}^{t}_{\mathrm{cVAE-DL}}+\mathcal{L}^{t}_{\mathrm{cLR-DL}}.
\end{split}
\end{equation}

\noindent
\textbf{Conflict Removal with Auxiliary Data.} Note that Equation~\ref{eqn:c1-conflict} contains conflicting objectives. The first term encourages the model to reconstruct the inputs of the current task, 
while the third term encourages the model to generate the same images as the outputs of the old model. In addition, the first term encourages the model to encode the input images to normal distributions, while the second term encourages the model to encode the input images to a distribution learned from the old model. Similar conflicting objectives exist in Equation~\ref{eqn:c2-conflict}. To sum up, the conflicts appear when the model is required to produce two different outputs, namely mimicking the performance of the old model and accomplishing the new goal, given the same inputs. 

To address these conflicting objectives, we propose to use auxiliary data for distilling knowledge from the old model $M_{t-1}$ to model $M_{t}$. The use of auxiliary data for distillation removes these conflicts.  It is important that new auxiliary data should be used for each task, otherwise the network could potentially implicitly encode them when learning previous tasks.  We describe approaches for doing so without requiring external data sources in Sec.~\ref{sec:auxiliary}.

The auxiliary data  $\mathbb{S}_t^{\mathrm{aux}}=\{(\mathbf{A}_{i,t}^{\mathrm{aux}}, \mathbf{B}_{i,t}^{\mathrm{aux}})|\mathbf{A}_{i,t}^{\mathrm{aux}} \in \mathbb{A}_t^{\mathrm{aux}}, \mathbf{B}_{i,t}^{\mathrm{aux}}\in{}\mathbb{B}_t^{\mathrm{aux}}\}_{i=1}^{N_t}$ consist of $N^{\mathrm{aux}}_t$ training pairs where $\mathbb{A}^{\mathrm{aux}}_t$ and $\mathbb{B}^{\mathrm{aux}}_t$ denote the domain of auxiliary conditional data and ground truth data respectively. For simplicity, we use the notations $\mathbf{A}^{\mathrm{aux}}_t, \mathbf{B}^{\mathrm{aux}}_t$ for an instance from the respective domain. 



The losses $\mathcal{L}^{t}_{\mathrm{cVAE-DL}}$ and $\mathcal{L}^{t}_{\mathrm{cLR-DL}}$ are re-written as:
\begin{equation}
\begin{split}
    & \mathcal{L}_{\mathrm{cVAE-DL}}^{t} = \mathcal{L}_{\mathrm{cVAE-GAN}}^{t} \\
    & + \beta \mathds{E}_{\mathbf{A}_{t}^{\mathrm{aux}},\mathbf{B}_{t}^{\mathrm{aux}} \sim p(\mathbf{A}_{t}^{\mathrm{aux}},\mathbf{B}_{t}^{\mathrm{aux}})} \, [ ||E_{t}(\mathbf{B}_{t}^{\mathrm{aux}})-E_{t-1}(\mathbf{B}_{t}^{\mathrm{aux}})||_{1}\\
    & + ||G_{t}(\mathbf{A}_{t}^{\mathrm{aux}},E_{t}(\mathbf{B}_{t}^{\mathrm{aux}})) -  G_{t-1}(\mathbf{A}_{t}^{\mathrm{aux}},E_{t-1}(\mathbf{B}_{t}^{\mathrm{aux}}))||_{1} ],
\end{split}
\label{eqn:c1}
\end{equation}

\vspace{-0.5cm}
\begin{equation}
\begin{split}
    & \mathcal{L}_{\mathrm{cLR-DL}}^{t} = \mathcal{L}_{\mathrm{cLR-GAN}}^{t} \\
    & + \beta \mathds{E}_{\mathbf{A}_{t}^{\mathrm{aux}} \sim p(\mathbf{A}_{t}^{\mathrm{aux}}),\mathbf{z} \sim p(\mathbf{z})} \, [ ||G_{t}(\mathbf{A}_{t}^{\mathrm{aux}},\mathbf{z})-G_{t-1}(\mathbf{A}_{t}^{\mathrm{aux}},\mathbf{z})||_{1} \\
    & + ||E_{t}(G_{t}(\mathbf{A}_{t}^{\mathrm{aux}},\mathbf{z}))-E_{t-1}(G_{t-1}(\mathbf{A}_{t}^{\mathrm{aux}},\mathbf{z}))||_{1}],\\
\end{split}
\label{eqn:c2}
\end{equation}
where $\beta$ is the loss weight for knowledge distillation.


Lifelong GAN can be used for continual learning of both image-conditioned and label-conditioned generation tasks. The auxiliary images for knowledge distillation for both settings can be generated using the Montage and Swap operations described in Section~\ref{sec:auxiliary}. For label-conditioned generation, we can simply use the categorical codes from previous tasks.  

\subsection{Auxiliary Data Generation}
\label{sec:auxiliary}

We now discuss the generation of auxiliary data. Recall from Section \ref{sec:i-kd} that we use auxiliary data to address the conflicting objectives in Equations \ref{eqn:c1-conflict} and \ref{eqn:c2-conflict}.


The auxiliary images do not require labels, and can in principle be sourced from online image repositories. However, this solution may not be scalable as it requires a new set of auxiliary images to be collected when learning each new task. A more desirable alternative may be to generate auxiliary data by using the current data in a way that avoids the over-fitting problem. We propose two operations for generating auxiliary data from the current task data:

\begin{enumerate}
\item Montage: Randomly sample small image patches from current input images and montage them together to produce auxiliary images for distillation. 
\item Swap: Swap the conditional image $\mathbf{A}_{t}$ and the ground truth image $\mathbf{B}_{t}$ for distillation. Namely the encoder receives the conditional image $\mathbf{A}_{t}$ and encodes it to a latent code $\mathbf{\widetilde{z}}$, and the generator is conditioned on the ground truth image $\mathbf{B}_{t}$.
\end{enumerate}

\noindent Both operations are used in image-conditioned generation; 
in label-conditioned generation, since there is no conditional image, only the montage operation is applicable.

Other alternatives may be possible. Essentially, the auxiliary data generation needs to provide out-of-task samples that can be used to preserve the knowledge learned by the old model. The knowledge is preserved using the distillation losses, which encourage the old and new models to produce similar responses on the out-of-task samples.
\vspace{-0.2cm}
\section{Experiments}

We evaluate Lifelong GAN for two settings: (1) image-conditioned image generation, and (2) label-conditioned image generation. We are the first to explore continual learning for image-conditioned image generation; no existing approaches are applicable for comparison. Additionally, we compare our model with the memory replay based approach which is the state-of-the-art for label-conditioned image generation.

\vspace{0.1cm}
\noindent
\textbf{Training Details.} All the sequential digit generation models are trained on images of size $64 \times 64$ and all other models are trained on images of size $128 \times 128$. We use the Tensorflow~\cite{abadi2016tensorflow} framework with Adam Optimizer~\cite{kingma2014adam} and a learning rate of $0.0001$. We set the parameters $\lambda_{\mathrm{latent}}=0.5$, $\lambda_{\mathrm{KL}}=0.01$, and $\beta=5.0$ for all experiments. The weights of generator and encoder in \textit{cVAE-GAN} and \textit{cLR-GAN} are shared. Extra training iterations on the generator and encoder using only distillation loss are used for models trained on images of size $128 \times 128$ for better remembering previous tasks.

\vspace{0.1cm}
\noindent
\textbf{Baseline Models.} We compare Lifelong GAN to the following baseline models: (a) \textit{Memory Replay (MR)}: Images generated by a generator trained on previous tasks are combined with the training images for the current task to form a hybrid training set. (b) \textit{Sequential Fine-tuning (SFT)}: The model is fine-tuned in a sequential manner, with parameters initialized from the model trained/fine-tuned on the previous task. (c) \textit{Joint Learning (JL)}: The model is trained utilizing data from all tasks. 

Note that for image-conditioned image generation, we only compare with joint learning and sequential fine-tuning methods, as memory replay based approaches are not applicable without any ground-truth conditional input.

\begin{figure}[ht]
\begin{center}
  \includegraphics[width=0.49\textwidth]{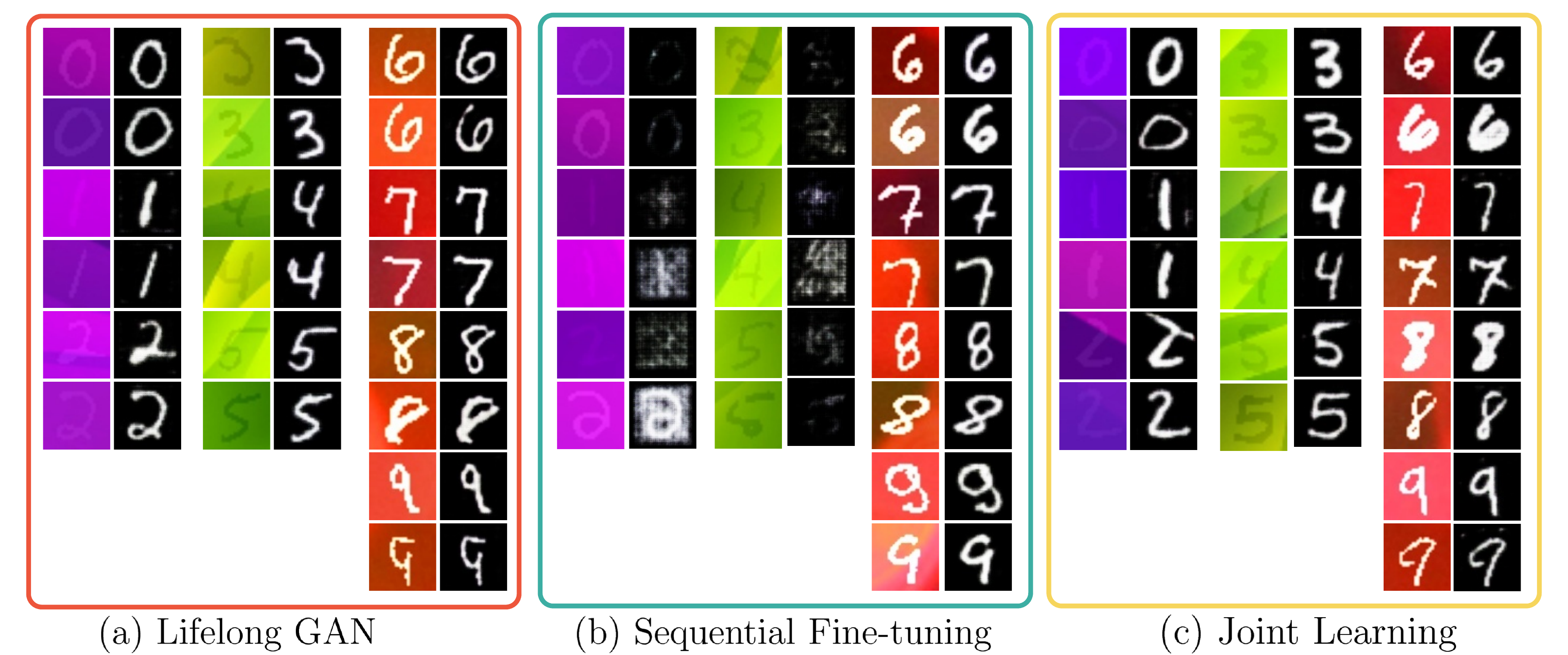}
\end{center}
\vspace{-0.6cm}
\caption{Comparison among different approaches for continual learning of MNIST digit segmentations. Lifelong GAN can learn the current task without forgetting the previous ones.}
\label{fig:mnist-seg}
\end{figure}

\begin{table}[ht]
\centering
\begin{tabular}{ c c c c c }
\toprule
& & SFT & JL & Ours \\
\cmidrule[0.04em](lr{.2em}){1-5} 
& Acc & 58.02 & 94.25 & 95.90 \\ 
MNIST & r-Acc & 61.56 & 96.79 & 96.14 \\  
& LPIPS & - & 0.150 & 0.157 \\
\cmidrule[0.04em](lr{.2em}){1-5} 
& Acc & 39.72 & 99.26 & 98.93 \\ 
Image-to-Image & r-Acc & 49.88 & 98.98 & 99.37 \\  
& LPIPS & - & 0.442 & 0.417 \\
\bottomrule
\end{tabular}
\vspace{-0.05in}
\caption{Quantitative evaluation for image-conditioned generation. For MNIST digit generation, LPIPS for real images is 0.154. For image-to-image translation, LPIPS for real images is 0.472.}
\vspace{-0.5cm}
\label{table:img2img}
\end{table}

\vspace{0.1cm}
\noindent 
\textbf{Quantitative Metrics.} We use different metrics to evaluate different aspects of the generation.
In this work, we use \textit{Acc}, \textit{r-Acc} and \textit{LPIPS} to validate the quality of the generated data. \textit{Acc} is the accuracy of the classifier network trained on real images and evaluated on generated images (higher indicates better generation quality). \textit{r-Acc} is the accuracy of the classifier network trained on generated  images and evaluated on real images (higher indicates better generation quality). \textit{LPIPS}~\cite{zhang2018unreasonable} is used to quantitatively evaluate the diversity as used in BicycleGAN~\cite{zhu2017toward}. Higher LPIPS indicates higher diversity. Furthermore, LPIPS closer to the ones of real images indicates more realistic generation.

\subsection{Image-conditioned Image Generation}

\noindent
\textbf{Digit Generation.}
We divide the digits in MNIST \cite{mnistlecun1998} into 3 groups: \{0,1,2\}, \{3,4,5\}, and \{6,7,8,9\}\footnote{group \{a,b,c\} contains digits with label a, b and c. This applies to all experiments on MNIST.}. The digits in each group are dyed with a signature color as shown in Figure~\ref{fig:mnist-seg}. Given a dyed image, the task is to generate a foreground segmentation mask for the digit (i.e. generate a foreground segmentation given a dyed image as condition). The three groups give us three tasks for sequential learning.

Generated images from the last task for all approaches are shown in Figure~\ref{fig:mnist-seg}. We can see that sequential fine-tuning suffers from catastrophic forgetting (it is unable to segment digits 0-5 from the previous tasks), while our approach can learn to generate segmentation masks for the current task without forgetting the previous tasks.

\noindent
\textbf{Image-to-image Translation.}
We also apply Lifelong GAN to more challenging domains and datasets with large variation for higher resolution images. 
The first task is image-to-image translation of edges $\rightarrow$ shoes photos~\cite{finegrained,xie15hed}. The second task is image-to-image translation of segmentations $\rightarrow$ facades~\cite{Tylecek13}. 
The goal of this experiment is to learn the task of semantic segmentations $\rightarrow$ facades without forgetting the task edges $\rightarrow$ shoe photos.
We sample {\raise.17ex\hbox{$\scriptstyle\mathtt{\sim}$}}20000 image pairs for the first task and use all images for the second task.

\begin{figure*}[h!]
 \centering
  \includegraphics[width=0.9\linewidth]{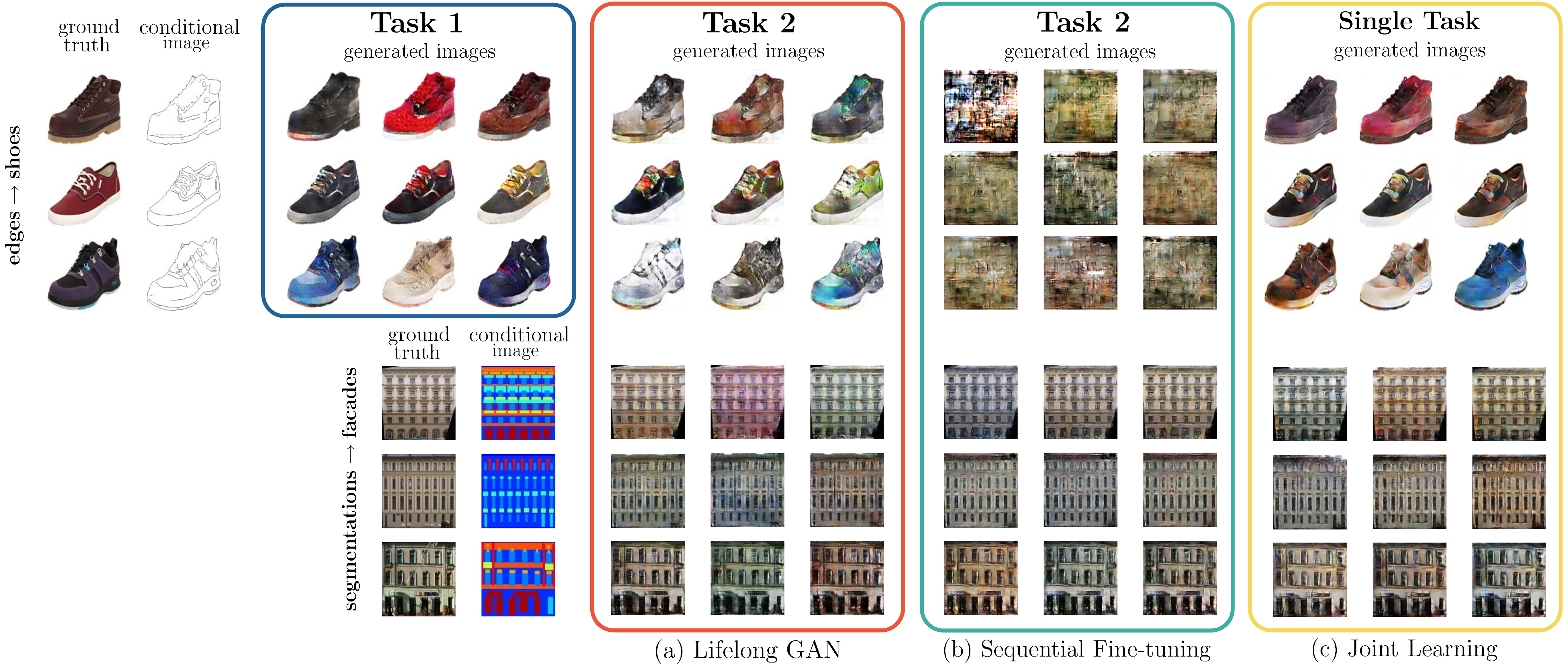}
\vspace{-0.35cm}
\caption{Comparison among different approaches for continual learning of image to image translation tasks. Given the same model trained for the task \textit{edges $\rightarrow$ shoes}, we train Lifelong GAN and sequential fine-tuning model on the task \textit{segmentations $\rightarrow$ facades}. Sequential fine-tuning suffers from severe catastrophic forgetting. In contrast, Lifelong GAN can learn the current task while remembering the old task.}
\vspace{-0.2cm}
\label{fig:img2img}
\end{figure*}

Generated images for all approaches are shown in Figure~\ref{fig:img2img}. For both Lifelong GAN and sequential fine-tuning, the model of \textit{Task2} is initialized from the same model trained on \textit{Task1}. We show the generation results of each task for Lifelong GAN. For sequential fine-tuning, we show the generation results of the last task. It is clear that the sequentially fine-tuned model completely forgets the previous task and can only generate incoherent facade-like patterns. In contrast, Lifelong GAN learns the current generative task while remembering the previous task. It is also observed that Lifelong GAN is capable of maintaining the diversity of generated images of the previous task. 

\begin{figure*}[h!]
\begin{center}
  \includegraphics[width=0.9\textwidth]{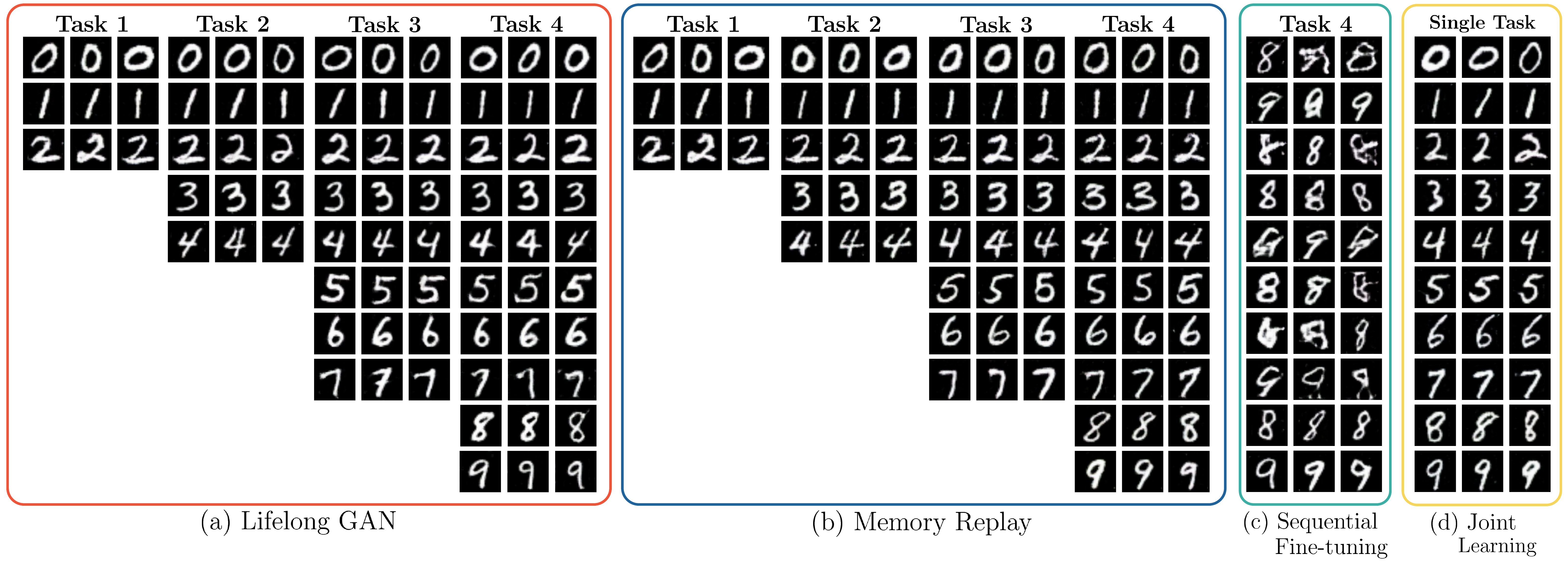}
\end{center}
\vspace{-0.7cm}
\caption{Comparison among different approaches for continual learning of MNIST digit generation conditioned on label. We demonstrate some intermediate results during different tasks of continual learning for our distillation based approach and memory replay. Sequential fine-tuning suffers from severe forgetting issues while other methods give visually similar results compared to the joint learning results.}
\vspace{-0.5cm}
\label{fig:mnist2}
\end{figure*}

\begin{table}[t]
\centering
\begin{tabular}{c c c c c c }
\toprule
 & & SFT & JL & MR & Ours \\
\cmidrule[0.04em](lr{.2em}){1-6} 
& Acc & 21.59 & 98.08 & 97.54 & 97.52 \\ 
MNIST & r-Acc & 21.21 & 87.72 & 85.57 & 87.77 \\ 
& LPIPS & - & 0.125 & 0.120 & 0.119 \\
\cmidrule[0.04em](lr{.2em}){1-6} 
& Acc & 20.0 & 96.4 & 87.6 & 98.4 \\ 
Flower & r-Acc & 19.6 & 83.6 & 60.4 & 85.6 \\ 
& LPIPS & - & 0.413 & 0.319 & 0.399  \\ 
\bottomrule
\end{tabular}
\vspace{-0.05in}
\caption{Quantitative evaluation for label-conditioned image generation tasks. For MNIST digit generation, LPIPS for real images is 0.155. For flower image generation, LPIPS for real images is 0.479.}
\vspace{-0.5cm}
\label{table:image-condition}
\end{table}

\begin{figure*}[h!]
\centering
  \includegraphics[width=0.75\linewidth]{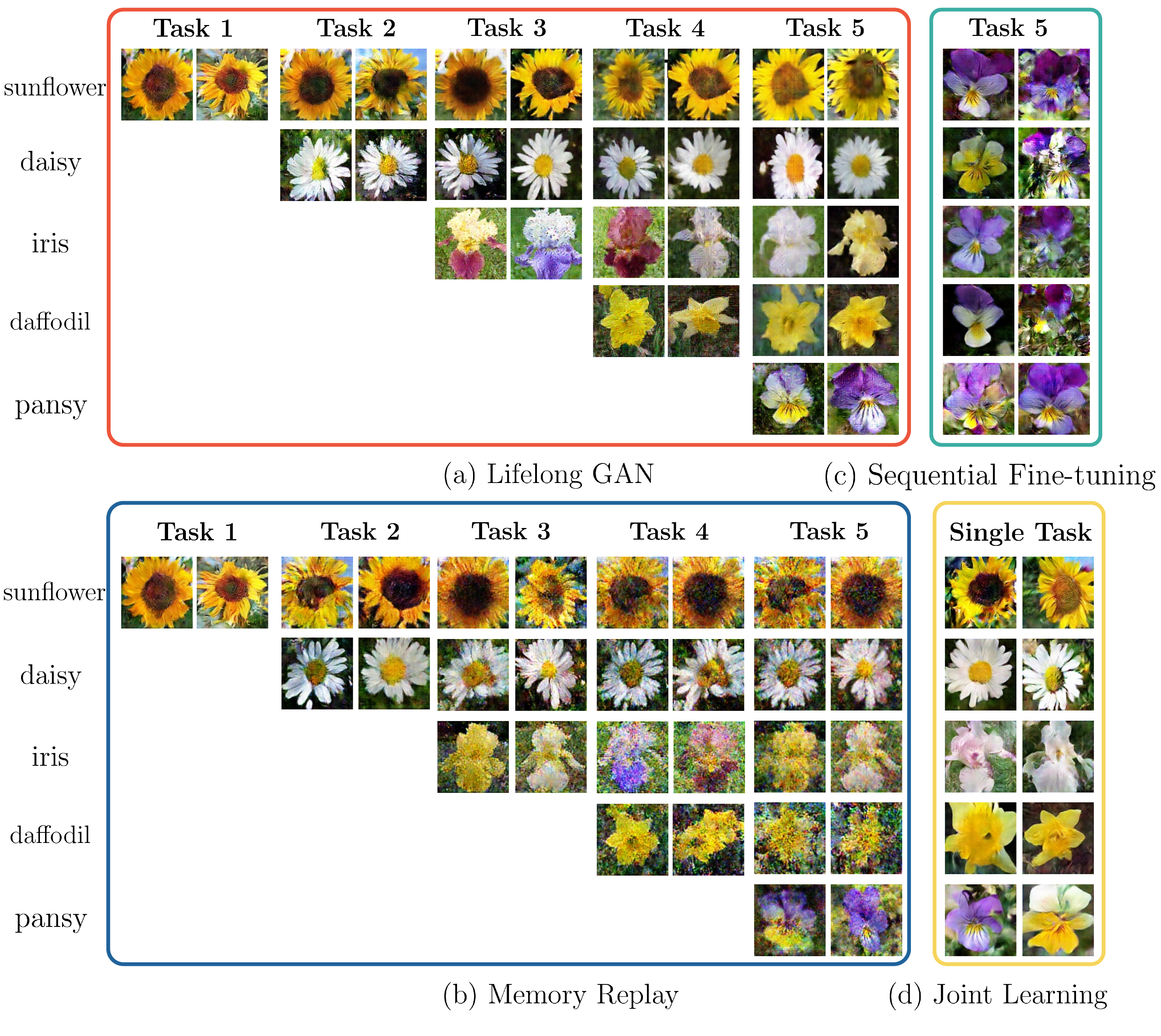}
\vspace{-0.6cm}
\caption{Comparison among different approaches for continual learning of flower image generation tasks. Given the same model trained for category \textit{sunflower}, we train Lifelong GAN, memory replay and sequential fine-tuning for other tasks. Sequential fine-tuning suffers from severe catastrophic forgetting, while both Lifelong GAN and memory replay can learn to perform the current task while remembering the old tasks. Lifelong GAN is more robust to artifacts in the generated images of the middle tasks, while memory replay is much more sensitive and all later tasks are severely impacted by these artifacts.}
\vspace{-0.5cm}
\label{fig:flower}
\end{figure*}

We conduct an \textit{ablation study} on image-to-image translation tasks. As per~\cite{zhu2017toward}, a system which stays faithful to the input should not exceed the LPIPS of real images. As shown in Tab.~\ref{table:img2img},  \textit{montage} and \textit{swap} improve performance.

\begin{table}[ht]
\centering
\vspace{-0.35cm}
\setstretch{0.6}
\setlength\extrarowheight{-5pt}
\begin{tabular}{ c c c c }
\toprule
\setlength\extrarowheight{-5pt}
& Ours & \makecell{w/o montage \\ w/o swap} & w/o  swap \\
\cmidrule[0.04em](lr{.2em}){1-4} 
Acc & 98.93 & 66.78 & 97.62 \\ 
r-Acc & 99.37 & 59.76 & 86.80 \\  
LPIPS & 0.417 & 0.518 & 0.490 \\
\bottomrule
\end{tabular}
\vspace{-0.35cm}
\caption{Ablation Study. LPIPS for real images is 0.472.}
\label{table:img2img}
\vspace{-5mm}
\end{table}


\subsection{Label-conditioned Image Generation}

\noindent
\textbf{Digit Generation.} 
We divide the MNIST \cite{mnistlecun1998} digits into 4 groups, \{0,1,2\}, \{3,4\}, \{5,6,7\} and \{8,9\}, resulting in four tasks for sequential learning. Each task is to generate binary MNIST digits given labels (one-hot encoded labels) as conditional inputs.

Visual results for all methods are shown in Figure~\ref{fig:mnist2}, where we also include outputs of the generator after each task for our approach and memory replay. Sequential fine-tuning results in catastrophic forgetting, as shown by this baseline's inability to generate digits from any previous tasks; when given a previous label, it will either generate something similar to the current task or simply unrecognizable patterns. Meanwhile, both our approach and memory replay are visually similar to joint training results, indicating that both are able to address the forgetting issue in this task. Quantitatively, our method achieves comparable classification accuracy to memory replay, and outperforms memory replay in terms of reverse classification accuracy.

\vspace{-0.5cm}

\paragraph{Flower Generation.}
We also demonstrate Lifelong GAN on a more challenging dataset, which contains higher resolution images from five categories of the Flower dataset~\cite{Nilsback06}. The experiment consists of a sequence of five tasks in the order of \textit{sunflower}, \textit{daisy}, \textit{iris}, \textit{daffodil},  \textit{pansy}. Each task involves learning a new category. 

Generated images for all approaches are shown in Figure~\ref{fig:flower}. We show the generation results of each task for both Lifelong GAN and memory replay to better analyze these two methods. For sequential fine-tuning, we show the generation results of the last task which is enough to show that the model suffers from catastrophic forgetting. 

Figure~\ref{fig:flower} gives useful insights into the comparison between Lifelong GAN and memory replay. Both methods can learn to generate images for new tasks while remembering previous ones. However, 
memory replay is more sensitive to generation artifacts appearing in the intermediate tasks of sequential learning. While training \textit{Task3} (category iris), both Lifelong GAN and memory replay show some artifacts in the generated images. For memory replay, the artifacts are reinforced during the training of later tasks and gradually spread over all categories. In contrast, Lifelong GAN is more robust to the artifacts and later tasks are much less sensitive to intermediate tasks.
Lifelong GAN treats previous tasks and current tasks separately, trying to learn the distribution of new tasks while mimicking the distribution of the old tasks.

\vspace{-0.1cm}
Table~\ref{table:image-condition} shows the quantitative results. Lifelong GAN outperforms memory replay by 10\% in terms of classification accuracy and 25\% in terms of reverse classification accuracy. 
We also observed visually and quantitatively that memory replay tends to lose diversity during the sequential learning, and generates images with little diversity for the final task. 

Moreover, to evaluate the quality of generated images, we conduct a user study with 20 participants. Each participant is given 60 image pairs (ours, baseline). For each pair, a participant is asked to pick the visually better image. Table~\ref{table:user} shows the percentage of pairs where ours are preferred. The user study indicates that our approach outperforms memory replay (MR) though is not on par with joint learning (JL). Note that we generate images from all categories to conduct the user study, thus it is not fair to include the sequential fine-tuning in the comparison as it forgets all previous tasks and generates images only for the last task.

\begin{table}[ht]
\vspace{0.2cm}
\setstretch{0.55}
\centering
\begin{tabular}{ c c c }
\toprule
flower (vs. MR) & flower (vs. JL) & img-to-img (vs. JL) \\
\cmidrule[0.04em](lr{.2em}){1-3} 
91.4\% & 28.2\% & 27.5\% \\ 
\bottomrule
\end{tabular}
\vspace{-0.3cm}
\caption{User Study.}
\label{table:user}
\vspace{-0.9cm}
\end{table}

\vspace{-0.2cm}
\section{Conclusion}
\vspace{-0.4cm}
We study the problem of lifelong learning for generative networks and propose a distillation based continual learning framework enabling a single network to be extended to new tasks without forgetting previous tasks with only supervision for the current task. Unlike previous methods that adopt memory replay to generate images from previous tasks as training data, we employ knowledge distillation to transfer learned knowledge from previous networks to the new network. Our generic framework enables a broader range of generation tasks including image-to-image translation, which is not possible using memory replay based methods. We validate Lifelong GAN for both image-conditioned and label-conditioned generation tasks, and both qualitative and quantitative results illustrate the generality and effectiveness of our method.


{\small
\bibliographystyle{ieee_fullname}
\bibliography{egbib}
}

\end{document}